\documentclass[conference]{IEEEtran}
\IEEEoverridecommandlockouts

\usepackage{cite}
\usepackage{amsmath,amssymb,amsfonts}
\usepackage{algorithmic}
\usepackage{graphicx}
\usepackage{textcomp}
\usepackage{xcolor}
\usepackage{multicol} %表格合并列
\usepackage{multirow} %表格合并行
\usepackage{booktabs} %表格边框线
\usepackage{colortbl} %表格颜色
\usepackage{subfig} %子图
\usepackage{adjustbox}
\usepackage{xspace}
\def\BibTeX{{\rm B\kern-.05em{\sc i\kern-.025em b}\kern-.08em
    T\kern-.1667em\lower.7ex\hbox{E}\kern-.125emX}}
    
\begin{document}
\newcommand{\ourapproach}{\textsc{Group-On}\xspace}
\newcommand{\ourmodule}{MoME\xspace}

\title{\ourapproach: Boosting One-Shot Segmentation with Supportive Query}

\author{\IEEEauthorblockN{1\textsuperscript{st} Hanjing Zhou\thanks{\IEEEauthorrefmark{1} These authors contributed equally.}\IEEEauthorrefmark{1}}
\IEEEauthorblockA{
\textit{Zhejiang University}\\
zhj85393@gmail.com}
\and
\IEEEauthorblockN{2\textsuperscript{nd} Mingze Yin\IEEEauthorrefmark{1}}
\IEEEauthorblockA{
\textit{Zhejiang University}\\
mzyin256@gmail.com}
\and
\IEEEauthorblockN{3\textsuperscript{rd} Danny Chen}
\IEEEauthorblockA{
\textit{University of Notre Dame}\\ dchen@cse.nd.edu}
\and
\IEEEauthorblockN{4\textsuperscript{th} Jian Wu\thanks{\IEEEauthorrefmark{2} Corresponding authors}\IEEEauthorrefmark{2}}
\IEEEauthorblockA{\textit{Zhejiang University} \\
\textit{Zhejiang Key Laboratory of Medical Imaging Artificial Intelligence}\\
wujian2000@zju.edu.cn}
\and
\IEEEauthorblockN{5\textsuperscript{th} Jintai Chen\IEEEauthorrefmark{2}}
\IEEEauthorblockA{
\textit{HKUST(Guangzhou)}\\
jtchen721@gmail.com}
}
% \usepackage{bibentry}
% END REMOVE bibentry

% \begin{document}

\maketitle

\begin{abstract}
One-shot semantic segmentation aims to segment query images given only ONE annotated support image of the same class. This task is challenging because target objects in the support and query images can be largely different in appearance and pose (\textit{i.e.}, intra-class variation). Prior works suggested that incorporating more annotated support images in few-shot settings boosts performances but increases costs due to additional manual labeling. In this paper, we propose a novel and effective approach for ONE-shot semantic segmentation, called \ourapproach \footnote{https://github.com/diaoshaoyou/Group-On}, which packs multiple query images in batches for the benefit of mutual knowledge support within the same category. Specifically, after coarse segmentation masks of the batch of queries are predicted, query-mask pairs act as pseudo support data to enhance mask predictions mutually.
To effectively steer such process, we construct an innovative \ourmodule module, where a flexible number of mask experts are guided by a scene-driven router and work together to make comprehensive decisions, fully promoting mutual benefits of queries. Comprehensive experiments on three standard benchmarks show that, in the ONE-shot setting, \ourapproach significantly outperforms previous works by considerable margins. With only one annotated support image, \ourapproach can be even competitive with the counterparts using 5 annotated 
images. 
\end{abstract}

% \begin{IEEEkeywords}
% component, formatting, style, styling, insert
% \end{IEEEkeywords}
\begin{figure*}
    \centering
    \subfloat[]{
    \includegraphics[width=0.31\textwidth]{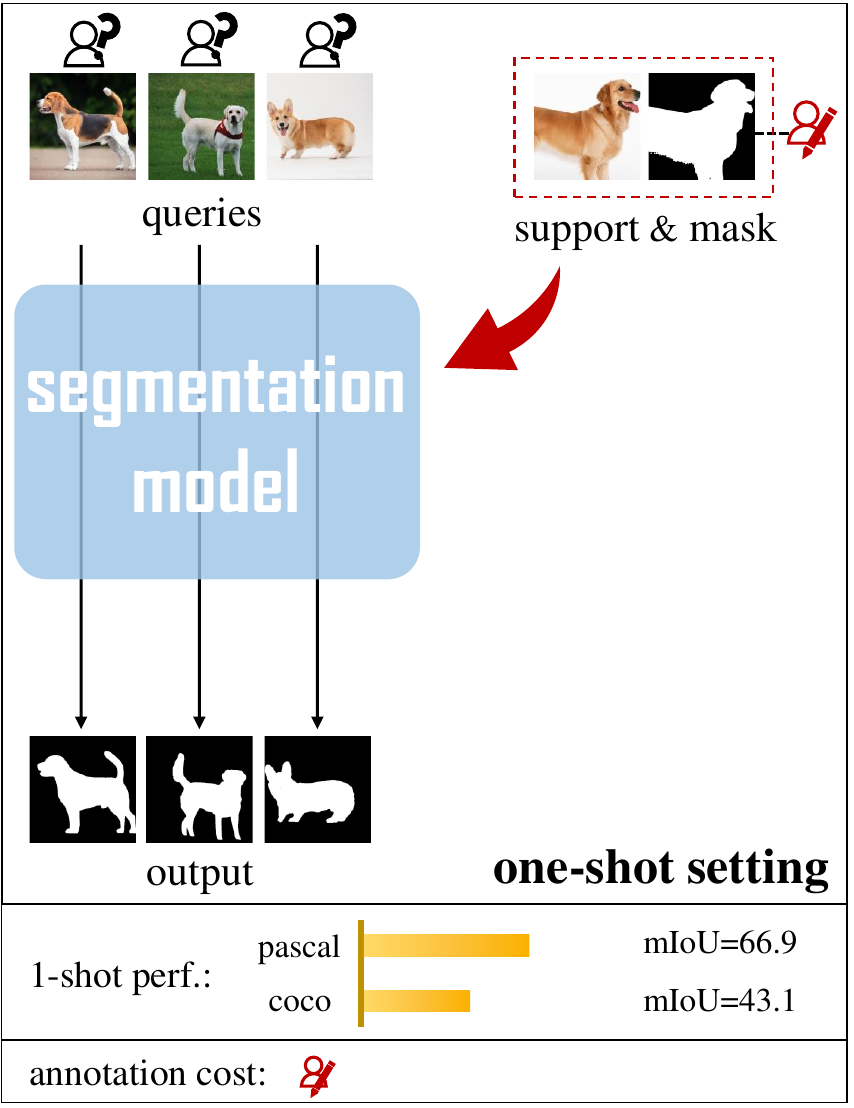}
    \label{Fig:1shot}
    }
    \hspace{-3mm}
    \subfloat[]{
    \includegraphics[width=0.31\textwidth]{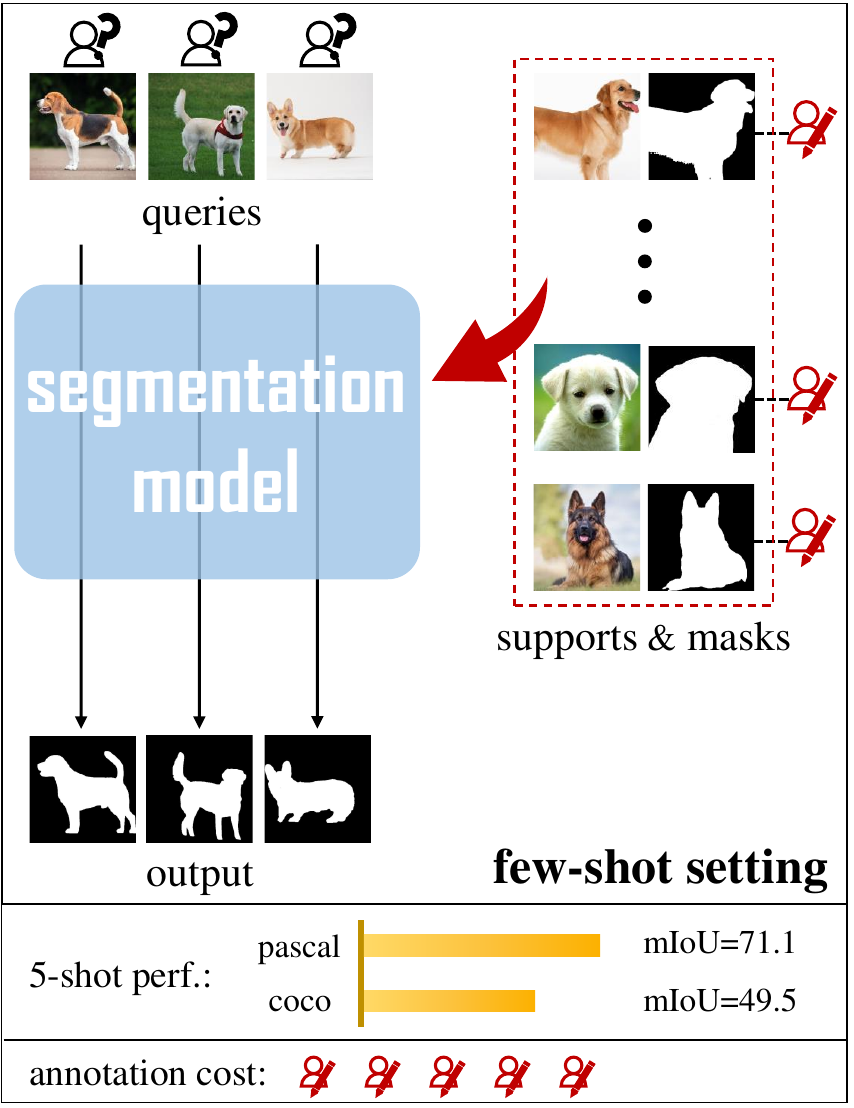}
    \label{Fig:fewshot}
    }
    \hspace{-2.2mm}
    \subfloat[]{
    \includegraphics[width=0.339\textwidth]{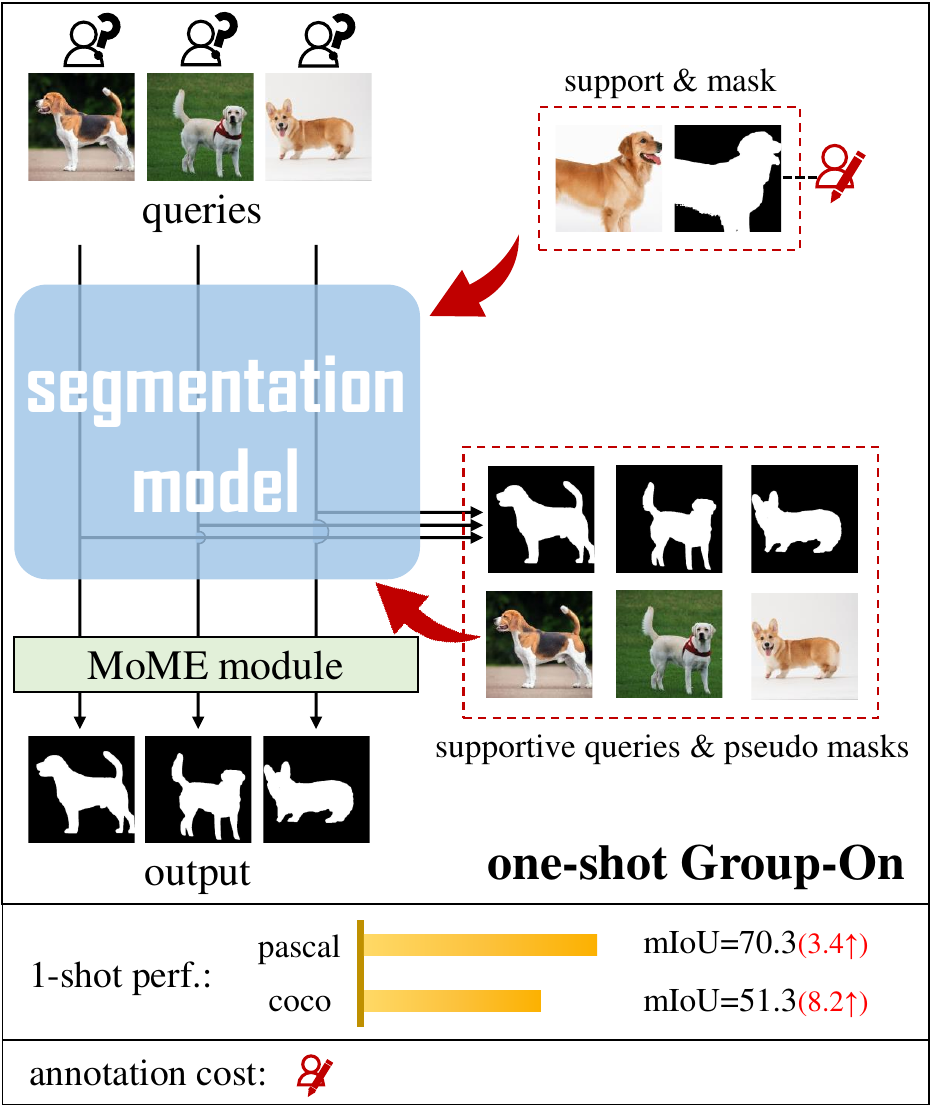}
    \label{Fig:brief}
    }\vspace{-0.5em}
    \caption{(a) To deal with queries from user, the common 1-shot setting uses one support image annotated by one annotator. (b) The common few-shot setting uses a few support images annotated by many extra annotators. (c) Our one-shot \ourapproach handles queries in bulk based on ONE labeled support without increasing extra annotators.
    All the queries are interchangeable and achieve mutual complementary benefits. \includegraphics[height=0.3cm, width=0.3cm]{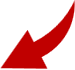} and \includegraphics[height=0.3cm, width=0.3cm]{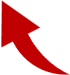} represent the input of support and pseudo support, respectively.
    The bar charts of segmentation performances on the bottom show that our one-shot \ourapproach even reaches a competitive level with the counterparts in 5-shot setting.}
    \vspace{-1em}
\label{Fig:1}
\end{figure*}
\section{Introduction}
\label{sec:intro}
Recently, few-shot learning has garnered significant attention due to its remarkable success in reducing the dependence on data annotations during the development of deep learning (DL) models. This paradigm offers an efficient way to deploy a DL model that can generalize well to new and unseen data by using only a sparing amount of category-identical reference data with manual labels. This is especially advantageous for tasks that require costly manual annotations, such as pixel-wise segmentation of images.
In a few-shot segmentation (FSS) practice, users require the segmentation model to segment unseen images (\textit{i.e.}, query) with guidance from one or several references (\textit{i.e.}, support) of the same class. Apparently, ``one-shot setting'' is the most annotation-effective solution, requiring the resource-intensive pixel-wise labeling of just one image. 
However, constrained by the limited information derived from just one manual annotation and the intrinsic diversity of the objects to be segmented, its performance often falls behind in other few-shot scenarios~\cite{ASnet}.

%介绍one-shot, few-shot能增强性能, few-shot缺点
In this paper, we rethink one-shot segmentation scenarios and find that its performance can reach levels comparable to counterparts under the few-shot segmentation setting.
Fig.~\ref{Fig:1}(a) shows traditional one-shot segmentation scenarios. A great deal of users submit image queries into the segmentation model. For each category, the model retrieves one support image identical to the queried category, accompanied by annotations from human annotators for segmentation. To boost performance, few-shot segmentation setting is introduced as in Fig.~\ref{Fig:1}(b), which uses more annotated support images. Although this augmentation in support data generally improves performance, it comes at the expense of requiring more costly manual annotations.
% few-shot更强的原因
Previous studies~\cite{Brinet, MuHS2023, FECAnet2023} attribute the superiority of the few-shot segmentation setting to its capability to overcome intra-class object variation, such as the diverse appearances and patterns within one category. The incorporation of additional supports exposes the model to a broader range of category-identical references, thereby introducing greater intra-class diversity, which typically leads to better performance.

Inspired by this, we try to integrate the merits of both one-shot and few-shot settings for fewer annotations and higher performances. Interestingly, a large quantity of query images from the same category can also offer intra-class multiplicity intuitively and thus provide mutual complementary support and benefits, which has been largely neglected in previous research yet. In this paper, we focus on one-shot segmentation that fixes the number of support images to only ONE. 
We propose a novel and effective approach for one-shot segmentation, called \ourapproach (Fig.~\ref{Fig:1}(c)), which simultaneously handles multiple queries input by users without increasing extra annotation costs and obtains better performance via mutual assistance among the queries.
For better understanding of \ourapproach, we define the query selected to be predicted as {\it host query}, while the rest queries as {\it supportive queries}. Supportive queries and their intermediate coarse masks, predicted from the only support image, serve as pseudo support data to facilitate segmentation of the host query. In other words, the host query is segmented by supports from one real support image and multiple pseudo ones (\textit{i.e.}, supportive queries), resulting in multiple predicted candidate masks. 
In pursuit of effective candidate mask integration, we construct an innovative \ourmodule module, where a scene-driven router steers a flexible number of mask experts towards jointly making comprehensive decisions on candidate masks. Such module consolidates expert decisions as the final result, expanding the model capability to deal with diverse images.

Note that the host and supportive queries hold an equal and interchangeable role. Each one among a group of query images can be the host query to be segmented finally, and all the query images offer favorable information mutually. Due to the need for testing many query samples in real-world scenarios, it is reasonable and feasible to collect and use various queries and apply \ourapproach for reciprocal help. Experiments on three standard benchmarks show that our approach significantly outperforms previous works by considerable margins, \textit{e.g.}, 8.2\% mIoU improvement on COCO-20$^i$~\cite{coco}. Our contributions are as follows:
\begin{itemize}
    \item
    We rethink the real-world application scenarios and propose \ourapproach, a novel and effective training approach for one-shot segmentation. For queries uploaded by users, \ourapproach processes category-identical ones together to address the intra-class variation issue by leveraging the rich object pattern variety. 
    \item
    We propose an innovative \ourmodule module for candidate mask integration, where a scene-driven router guides a flexible number of mask experts to cooperate and make decisions, expanding the model capability to handle various images.
    \item
    We perform extensive experiments and analysis on three standard and challenging FSS benchmarks. The results show that the performances of one-shot segmentation baselines with our \ourapproach are comparable to the counterparts of conventional 5-shot segmentation. 
\end{itemize}

\section{Methodology}
\label{sec:method}
\begin{figure*}[t]
    \centering
    \includegraphics[width=0.95\linewidth]{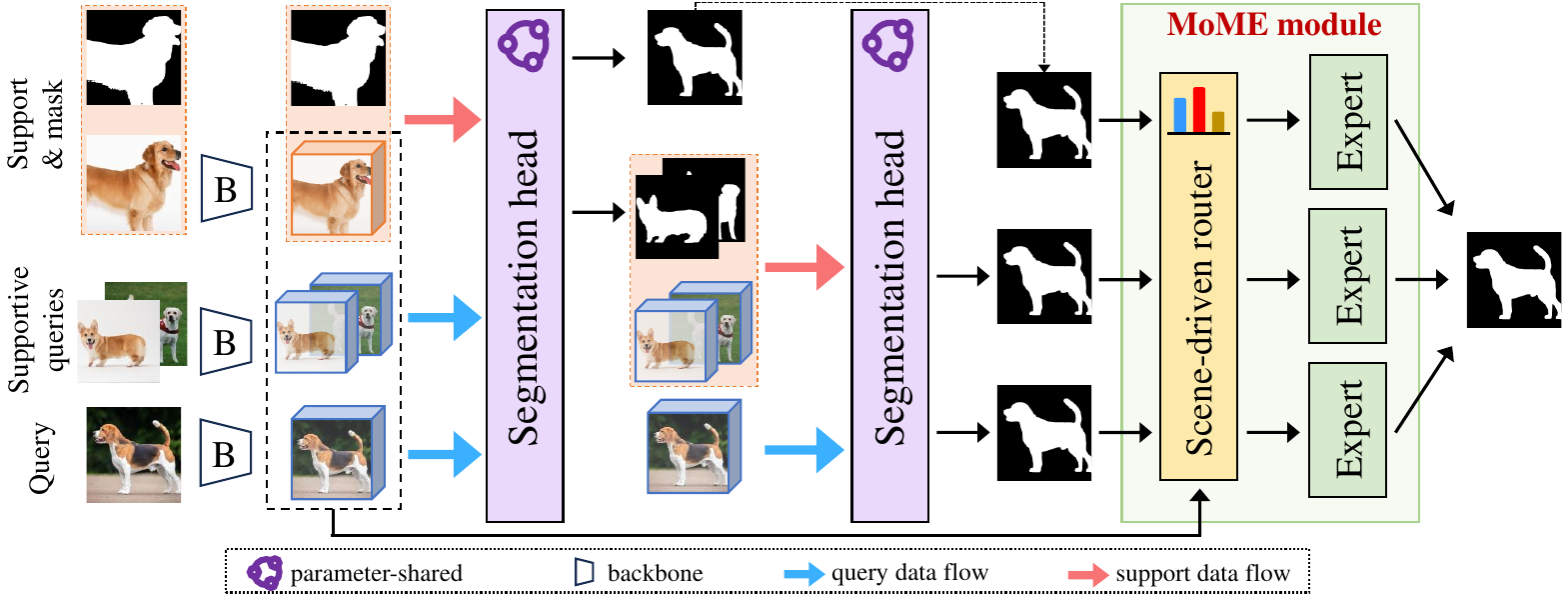}
    \caption{Illustrating \ourapproach in the one-shot setting. 
    Different from the conventional few-shot setting, a group of query images along with one support image is input into the model. After the coarse masks of all the query images are segmented, the supportive image-mask pairs function as pseudo support to segment the host query again. Eventually, the final result is produced by the proposed \ourmodule module, where a flexible number of mask experts make decisions on candidate masks guided by a scene-driven router. Note that the host and supportive queries play an equal and interchangeable role.}
    \label{Fig:overview}
    \vspace{-1em}
\end{figure*}
\subsection{Problem Setting}
%大体概括
%训练集和测试集
%每个episode中support set和query set 
%k-shot setting + 铺垫groupon设定
Few-shot segmentation (FSS) aims to segment images with limited annotated examples. Following prior works \cite{HSnet,VAT}, we adopt the prevalent meta-training paradigm called episodic training \cite{MatchNetwork}. Concretely, a dataset is divided into two class-disjoint sets $\mathcal{D}_{train}$ and $\mathcal{D}_{test}$, each of which contains many episodes. In an episode, there exist a support set $\mathcal{S}=\{(\textit{I}^\text{s}, \textit{M}^\text{s})\}$ and a query set $\mathcal{Q}=\{(\textit{I}^\text{q}, \textit{M}^\text{q})\}$ from the same class, where $\textit{I}$ and $\textit{M}$ represent raw images and their masks. Each episode instantiates an $N$-way $K$-shot segmentation task, which means selecting $K$ support image-label pairs per category from $N$ different categories to construct $\mathcal{S}$. Current approaches typically use one support image-label pair for training whereas one (\textit{i.e.}, 1-way 1-shot) or five (\textit{i.e.}, 1-way 5-shot) for testing, and one query image during both stages. The number of support images matters since it determines annotation burdens. We always seek to obtain similar performances via as few support images as possible and preferably only one. That is why many models \cite{OSLSM, SG-One} freeze the number of support images to one for training.

In this work, apart from the 1-way setting following \cite{HSnet}, we propose and undertake a 1-support $N_q$-query setting ($N_q>1$) extended from the conventional settings. That is, we use ONE support image but multiple query images to help each other in the process. 

\subsection{Group-On Process under One-Shot Setting}
%流程: 1.get pseudo support   2. get candidate masks 
%train & test的不同
Aiming to maximize the semantic information of different patterns in the same class as much as possible, we use a group of $N_q$ query images and one support image-label pair as input to the model. The pivotal competitive advantage of \ourapproach (see Fig. 2) is in utilizing the group of query images for mutual knowledge complement.
Among the $N_q$ query images in the group, the final predicted image is called the host query while the rest images are supportive queries. For simple illustration, in this section we assume that $N_q=2$, and the host query and its feature are denoted as $\textit{I}^{\text{host}}$ and $\textit{F}^{\text{host}}$, while the supportive ones as $\textit{I}^{\text{supp}}$ and $\textit{F}^{\text{supp}}$. 

The entire data stream goes through the segmentation head $f$ twice. In the first time, the support feature-mask pair ($\textit{F}^{\text{s}}$, $\textit{M}^{\text{s}}$) and the query features $\textit{F}^{\text{host}},~\textit{F}^{\text{supp}}_i$'s are input into the model to generate the initial coarse masks, as: 
\begin{equation}
\setlength{\abovedisplayskip}{5pt}
\setlength{\belowdisplayskip}{5pt}
\label{Eq:1}
\textit{\textbf{C}}_{1}=f(\textit{F}^{\text{s}},~\textit{M}^{\text{s}},~\textit{F}^{\text{host}}), \ \ \ \ \ 
% \end{equation}
% \begin{equation}
% \setlength{\abovedisplayskip}{1pt}
% \setlength{\belowdisplayskip}{1pt}
% \label{Eq:2}
\hat{\textit{M}}^{\text{supp}}_i=f(\textit{F}^{\text{s}},~\textit{M}^{\text{s}},~\textit{F}^{\text{supp}}_{i}),
\end{equation}
where $\textit{\textbf{C}}_{1}$ is a candidate mask for the host query, and $\hat{\textit{M}}^{\text{supp}}_i$ is the coarse mask predicted using the real support feature and its mask ($\textit{F}^{\text{s}}$, $\textit{M}^{\text{s}}$), $i=1,\ldots,N_q-1$ ($N_q=2$ here).

Those predicted coarse masks $\hat{\textit{M}}^{\text{supp}}_1$, $\ldots,$ $\hat{\textit{M}}^{\text{supp}}_{N_q-1}$ and their corresponding supportive queries are used as pseudo support data, which are input into the segmentation head again to compute the other candidate masks, as:
\begin{equation}
\setlength{\abovedisplayskip}{8pt}
\setlength{\belowdisplayskip}{8pt}
\textit{\textbf{C}}_{j+1}=f(\textit{F}^{\text{supp}}_{j},~\hat{\textit{M}}^{\text{supp}}_j,~\textit{F}^{\text{host}}), \ \ j=1,\ldots,N_q-1, 
\label{Eq:3}
\end{equation}
where $\textit{\textbf{C}}_{j+1}$ is the candidate mask of $\textit{I}^{\text{host}}$ derived from the supportive feature-mask pair ($\textit{F}^{\text{supp}}_{j},~\hat{\textit{M}}^{\text{supp}}_j$), $j=1,\ldots,N_q-1$.
Hence, we obtain $N_q$ candidate masks $\textit{\textbf{C}}_1, \ldots, \textit{\textbf{C}}_{N_q}$, produced from true and pseudo references in distinct appearances but from the same class, thus increasing the object diversity. The final result $\hat{\textit{M}}^{\text{host}}$ is computed by fusing these $N_q$ candidate masks via a novel \ourmodule module ($\texttt{MoME}$), as:
\begin{equation}
\setlength{\abovedisplayskip}{5pt}
\setlength{\belowdisplayskip}{5pt}
\label{Eq:4}
\hat{\textit{M}}^{\text{host}
}=\texttt{MoME}(\textit{\textbf{C}}_{1},
    \ldots, \textit{\textbf{C}}_{N_q}).
\end{equation}
It is noteworthy that under the 1-shot setting, all the query images are treated equivalently when applying \ourapproach. That is, the host query is not limited only to $\textit{I}^{\text{host}}$ but must be every query image, and the remaining supportive queries are processed in the same way as what is described above. This process is able to attain mutual knowledge complement. The reason for this is that it allows to make the most out of clues hidden in the query data and promote efficiency.
\begin{table*}[ht]
    \caption{Segmentation performances on the COCO-20$^i$ (left) and PASCAL-5$^i$ (right) datasets in the 1-support $N_q$-query setting. HSNet$^5$ and ASNet$^5$ represent results of the baselines HSNet (HS) and ASNet (AS) under the conventional 5-support setting, while the rest results are all in the 1-support setting. $^\dagger$HSNet, $^\dagger$ASNet: Our reimplementation with the CLIP backbone.}\vspace{-1em}
    % based on official codes. We highlight the best results in \textbf{bold} and \underline{underline} the second best results for better viewing.}\vspace{-1em}
    \label{Tab:pascal_coco}
    \begin{center}
    \setlength{\tabcolsep}{0.2em}
    \resizebox{\textwidth}{!}{
    \begin{tabular}{c|l|c c c c c c|c c c c c c}
            \toprule
            \multirow{2}{*}{Setting} & \multicolumn{1}{c|}{\multirow{2}{*}{Method}} & \multicolumn{6}{c}{COCO-20$^i$~\cite{coco}} & \multicolumn{6}{|c}{PASCAL-5$^i$~\cite{OSLSM}} \\
            \cmidrule{3-14}
             & & Fold-0 & Fold-1 & Fold-2 & Fold-3 & Mean & FB-IoU & Fold-0 & Fold-1 & Fold-2 & Fold-3 & Mean & FB-IoU\\
            \midrule
            \multicolumn{14}{c}{\textbf{\textit{Model performances on the ResNet-50 backbone}}}\\
            \midrule
            \multirow{2}{*}{5-support} & HSNet$^5$ \cite{HSnet} & \textit{43.3} & \textit{51.3} & \textit{48.2} & \textit{45.0} & \textit{46.9} & \textit{70.7} &  \textit{70.3} & \textit{73.2} & \textit{67.3} & \textit{67.1} & \textit{69.5} & \textit{80.6}  \\
            & ASNet$^5$ \cite{ASnet} &  \textit{47.6} &  \textit{50.1} &  \textit{47.7} &  \textit{46.4} &  \textit{47.9} &  \textit{71.6} & \textit{72.6} & \textit{74.3} &  \textit{65.3} &  \textit{67.1} &  \textit{70.8} &  \textit{80.4}  \\
            \midrule
            \multirow{6}{*}{1-support}
            % DCAMA \cite{DCAMA} & 41.9 & 45.1 & 44.4 & 41.7 & 43.3 & 69.5 & 67.5 & 72.3 & 59.6 & 59.0 & 64.6 & 75.7 \\
            % &VAT \cite{VAT} & 39.0 & 43.8 & 42.6 & 39.7 & 41.3 & 68.8 & 67.6 & 71.2 & 62.3 & 60.1 & 65.3 & 77.4 \\
            % &FECANet \cite{FECAnet2023} & 38.5 & 44.6 & 42.6 & 40.7 & 41.6 & 69.6 & 69.2 & 72.3 & \underline{62.4} & \underline{65.7} & 67.4 & 78.7  \\
            % & QCLNet \cite{QCLnet2023} & 39.8 & 45.7 & 42.5 & 41.2 & 42.3 & - & 65.2 & 70.3 & 60.8 & 61.0 & 64.3 & -  \\
            % & HPA \cite{HPA} & 40.3 & 46.6 & 44.1 & 42.7 & 43.4 & 68.2 & 65.9 & 72.0 & 64.7 & 56.8 & 64.8 & 76.4  \\
            & ABCNet \cite{ABCNet} & \underline{42.3} & 46.2 & 46.0 & 42.0 & 44.1 & 69.9 & 68.8 & 73.4 & 62.3 & 59.5 & 66.0 & 76.0  \\
            & MCE \cite{MCE} & 42.1 & 48.3 & 43.7 & 42.76 & 44.2 & - & 65.3 & 71.2 & \textbf{66.2} & 61.0 & 65.9 & 78.1  \\
            % & RiFeNet \cite{RiFeNet} & 39.1 & 47.2 & 44.6 & 45.4 & 44.1 & - & 68.4 & 73.5 & 67.1 & 59.4 & 67.1 & - \\
            &HSNet \cite{HSnet} & 36.3 & 43.1 & 38.7 & 38.7 & 39.2 & 68.2 & 64.3 & 70.7 & 60.3 & 60.5 & 64.0 & 76.7  \\
            &ASNet \cite{ASnet} & 41.5 & 44.1 & 42.8 & 40.6 & 42.4 & 68.8 & 68.9 & 71.7 & 61.1 & 62.7 & 66.1 & 77.7 \\
            \cmidrule{2-14}
            &\ourapproach (HS) & \cellcolor{gray!20}41.9 & \cellcolor{gray!20}\underline{48.9} & \cellcolor{gray!20}\underline{46.7} & \cellcolor{gray!20}\underline{43.1} & \cellcolor{gray!20}\underline{45.2} (\textcolor{red}{\footnotesize{6.0}\scalebox{0.9}{$\uparrow$}}) & \cellcolor{gray!20}\underline{71.8} (\textcolor{red}{\footnotesize{3.6}\scalebox{0.9}{$\uparrow$}}) & \cellcolor{gray!20}\underline{71.3} & \cellcolor{gray!20}\underline{73.5} & \cellcolor{gray!20}61.8 & \cellcolor{gray!20}64.2 & \cellcolor{gray!20}\underline{67.7} (\textcolor{red}{\footnotesize{3.7}\scalebox{0.9}{$\uparrow$}}) & \cellcolor{gray!20}\underline{78.8} (\textcolor{red}{\footnotesize{2.1}\scalebox{0.9}{$\uparrow$}})  \\
            &\ourapproach (AS) & \cellcolor{gray!20}\textbf{46.4} & \cellcolor{gray!20}\textbf{49.6} & \cellcolor{gray!20}\textbf{48.6} & \cellcolor{gray!20}\textbf{46.7} & \cellcolor{gray!20}\textbf{47.8} (\textcolor{red}{\footnotesize{5.4}\scalebox{0.9}{$\uparrow$}}) & \cellcolor{gray!20}\textbf{72.6} (\textcolor{red}{\footnotesize{3.8}\scalebox{0.9}{$\uparrow$}}) & \cellcolor{gray!20}\textbf{73.1} & \cellcolor{gray!20}\textbf{73.9} & \cellcolor{gray!20}61.7 & \cellcolor{gray!20}\textbf{67.7} & \cellcolor{gray!20}\textbf{69.1} (\textcolor{red}{\footnotesize{3.0}\scalebox{0.9}{$\uparrow$}}) & \cellcolor{gray!20}\textbf{80.1} (\textcolor{red}{\footnotesize{2.4}\scalebox{0.9}{$\uparrow$}}) \\
            % resnet101
            \midrule
            \multicolumn{14}{c}{\textbf{\textit{Model performances on the ResNet-101 backbone}}}\\
            \midrule
            \multirow{2}{*}{5-support} & HSNet$^5$ \cite{HSnet} & \textit{45.9} & \textit{53.0} & \textit{51.8} & \textit{47.1} & \textit{49.5} & \textit{72.4} & \textit{71.8} & \textit{74.4} & \textit{67.0} & \textit{68.3} & \textit{70.4} & \textit{80.6}  \\
            & ASNet$^5$ \cite{ASnet}  & \textit{48.0} & \textit{52.1} & \textit{49.7} & \textit{48.2} & \textit{49.5} & \textit{72.7} & \textit{73.1} & \textit{75.6} & \textit{65.7} & \textit{69.9} & \textit{71.1} & \textit{81.0}  \\
            \midrule
            \multirow{6}{*}{1-support}  
            % & DCAMA \cite{DCAMA} & 41.5 & 46.2 & 45.2 & 41.3 & 43.5 & 69.9 &  65.4 & 71.4 & 63.2 & 58.3 & 64.6 & 77.6  \\
            % & VAT \cite{VAT}  & 39.5 & 44.4 & 46.1 & 40.4 & 42.6 & - & 68.4 & 72.5 & 64.8 & 64.2 & 67.5 & 78.8 \\
            % & HPA \cite{HPA} & 43.1 & 50.0 & 44.8 & 45.2 & 45.8 & 68.4 & 66.4 & 72.7 & 64.1 & 59.4 & 65.6 & 76.6 \\
            % & ABCNet \cite{ABCNet} & - & - & - & - & - & - & 65.3 & 72.9 & \underline{65.0} & 59.3 & 65.6 & 78.5 \\
            & QCLNet \cite{QCLnet2023} & 40.0 & 45.5 & 45.1 & 43.6 & 43.6 & - & 67.9 & 72.5 & 64.3 & 63.4 & 67.0 & -  \\
            & SCCAN \cite{SCCAN} & 42.6 & 51.4 & \underline{50.0} & \underline{48.8} & 48.2 & 69.7 & 70.0 & 73.9 & \textbf{66.8} & 61.7 & 68.3 & 78.5  \\
            % &TBSNet \cite{TBSNet} & - & - & - & - & - & - & 68.5 & 72.0 & 63.8 & 59.5 & 65.9 & 77.7 \\ 
            % & RiFeNet \cite{RiFeNet} & - & - & - & - & - & - & 68.9 & 73.8 & 66.2 & 60.3 & 67.3 & -\\
            & HSNet \cite{HSnet} & 37.2 & 44.1 & 42.4 & 41.3 & 41.2 & 69.1 & 67.3 & 72.3 & 62.0 & 63.1 & 66.2 & 77.6  \\
            & ASNet \cite{ASnet} & 41.8 & 45.4 & 43.2 & 41.9 & 43.1 & 69.4 & 69.0 & 73.1 & 62.0 & 63.6 & 66.9 & 78.0  \\
            \cmidrule{2-14}
            & \ourapproach (HS) & \cellcolor{gray!20}\underline{44.9} & \cellcolor{gray!20}\underline{53.4} & \cellcolor{gray!20}48.6 & \cellcolor{gray!20}47.7 & \cellcolor{gray!20}\underline{48.7} (\textcolor{red}{\footnotesize{7.5}\scalebox{0.9}{$\uparrow$}}) & \cellcolor{gray!20}\underline{73.4} (\textcolor{red}{\footnotesize{4.3}\scalebox{0.9}{$\uparrow$}})& \cellcolor{gray!20}\underline{72.6} & \cellcolor{gray!20}\underline{75.0} & \cellcolor{gray!20}63.5 & \cellcolor{gray!20}\underline{66.8} & \cellcolor{gray!20}\underline{69.5} (\textcolor{red}{\footnotesize{3.3}\scalebox{0.9}{$\uparrow$}}) & \cellcolor{gray!20}\textbf{80.4} (\textcolor{red}{\footnotesize{2.8}\scalebox{0.9}{$\uparrow$}}) \\
            & \ourapproach (AS) & \cellcolor{gray!20}\textbf{48.1} & \cellcolor{gray!20}\textbf{54.2} & \cellcolor{gray!20}\textbf{51.6} & \cellcolor{gray!20}\textbf{51.3} & \cellcolor{gray!20}\textbf{51.3} (\textcolor{red}{\footnotesize{8.2}\scalebox{0.9}{$\uparrow$}}) & \cellcolor{gray!20}\textbf{73.5} (\textcolor{red}{\footnotesize{4.1}\scalebox{0.9}{$\uparrow$}}) & \cellcolor{gray!20}\textbf{73.8} & \cellcolor{gray!20}\textbf{75.5} & \cellcolor{gray!20}62.5 & \cellcolor{gray!20}\textbf{69.3} & \cellcolor{gray!20}\textbf{70.3} (\textcolor{red}{\footnotesize{3.4}\scalebox{0.9}{$\uparrow$}}) & \cellcolor{gray!20}\underline{80.2} (\textcolor{red}{\footnotesize{2.2}\scalebox{0.9}{$\uparrow$}})  \\
            \midrule
            % clip
            \multicolumn{14}{c}{\textbf{\textit{Model performances on the CLIP backbone}}}\\
            \midrule
            \multirow{5}{*}{1-support} & PGMA-Net~\cite{PGMANet} & - & - & - & - & - & - & 74.0 & \textbf{81.9} & 66.8 & 73.7 & 74.1 & 82.1 \\
            & $^\dagger$HSNet~\cite{HSnet} & 53.4 & 60.5 & 58.3 & 57.3 & 57.4 & 75.0 & 72.5 & 76.4 & 68.6 & 69.6 & 71.8 & 82.6  \\
            & $^\dagger$ASNet~\cite{ASnet} & 53.6 & 60.8 & 57.8 & 56.4 & 57.1 & 75.1 & 74.4 & 77.8 & 68.3 & 73.2 & 73.4 & 82.8  \\
            \cmidrule{2-14}
            & \ourapproach (HS)  & \cellcolor{gray!20}\underline{54.5} & \cellcolor{gray!20}\textbf{62.8} & \cellcolor{gray!20}\underline{58.5} & \cellcolor{gray!20}\underline{57.5} & \cellcolor{gray!20}\underline{58.3} (\textcolor{red}{\footnotesize{0.9}\scalebox{0.9}{$\uparrow$}}) & \cellcolor{gray!20}\underline{78.1} (\textcolor{red}{\footnotesize{3.0}\scalebox{0.9}{$\uparrow$}}) & \cellcolor{gray!20}\underline{77.1} & \cellcolor{gray!20}\underline{80.5} & \cellcolor{gray!20}\textbf{68.8} & \cellcolor{gray!20}\underline{74.6} & \cellcolor{gray!20}\underline{75.3} (\textcolor{red}{\footnotesize{3.5}\scalebox{0.9}{$\uparrow$}}) & \cellcolor{gray!20}\textbf{84.7} (\textcolor{red}{\footnotesize{2.1}\scalebox{0.9}{$\uparrow$}})  \\
            & \ourapproach (AS) & \cellcolor{gray!20}\textbf{55.0} & \cellcolor{gray!20}\underline{61.8} & \cellcolor{gray!20}\textbf{59.0} & \cellcolor{gray!20}\textbf{57.9} & \cellcolor{gray!20}\textbf{58.4} (\textcolor{red}{\footnotesize{1.3}\scalebox{0.9}{$\uparrow$}}) & \cellcolor{gray!20}\textbf{78.2} (\textcolor{red}{\footnotesize{3.1}\scalebox{0.9}{$\uparrow$}}) & \cellcolor{gray!20}\textbf{77.9} & \cellcolor{gray!20}79.7 & \cellcolor{gray!20}\underline{68.3} & \cellcolor{gray!20}\textbf{76.9} & \cellcolor{gray!20}\textbf{75.7} (\textcolor{red}{\footnotesize{2.3}\scalebox{0.9}{$\uparrow$}}) & \cellcolor{gray!20}\underline{84.2} (\textcolor{red}{\footnotesize{1.4}\scalebox{0.9}{$\uparrow$}}) \\
            \bottomrule
    \end{tabular}}
    % \end{small}
    \end{center}
    \vspace{-2em}
\end{table*}
\subsection{Mixture of Mask Experts Module}
\label{sec:voting} 
%motivation
%强调one-shot/mutual support.  model up->coarse mask up->prediction up
For advanced organization and utilization of candidate masks, we propose a novel Mixture of Mask Experts (MoME) module inspired by MoE. Since candidate masks are predicted based on supportive queries/support image that contain intra-class variation, there is an urgent need for multiple mask experts proficient in different aspects to handle the candidate masks effectively. Since the support image and supportive queries function similarly in this module, they are collectively referred to as supportive query data below for simplicity.

As illustrated in Fig.~\ref{Fig:overview}, a scene-driven router evaluates and scores expert decisions based on query image scenes, with each mask expert responsible for one candidate and working together to overcome the intra-class variation challenge.
Firstly, a scene-driven router (\texttt{Router}) evaluates the contribution of each expert based on scene differences between host and supportive queries, 
generating router weights $W$ through cross-attention:
\begin{equation}
    W = \texttt{Router}(\textit{F}^{\text{host}}, \textit{F}^{\text{supp}}_{1}, \textit{F}^{\text{supp}}_{2}, \ldots, \textit{F}^{\text{supp}}_{N_q}),
\end{equation}
where $W=[w_1, \ldots, w_i, \ldots, w_{N_q}]$ and $w_i$ is the router weight of $i_{th}$ candidate mask.
Steered by the router, secondly we form $N_q$ decisions by feeding candidate masks into an ensemble of mask experts (\texttt{Expert}), each of which comprises a convolutional layer with different weights. 
Eventually, we obtain the elegant final result $\hat{\textit{M}}^{\text{host}}$ through a weighted sum of mask expert decisions:
\begin{equation}
    \hat{\textit{M}}^{\text{host}}= \sum_{i=1}^{N_q} \texttt{Expert}(\textit{\textbf{C}}_i)*w_{i}.
\end{equation}

With a flexible number of mask experts (\textit{i.e.}, query amount $N_q$), \ourmodule module facilitates more efficient integration of candidate masks and significantly expands the model capability to handle different types of images. Furthermore, such module provides robust technical assurance for mutual benefits of multiple queries in \ourapproach, alleviating severe intra-class variation issue.

\section{Experiments}
\label{sec:exp}
% \subsection{Setup}
\label{sec:setup}
\paragraph{Datasets and Metrics}
We evaluate \ourapproach on three standard benchmarks: COCO-$20^i$ \cite{coco}, PASCAL-$5^i$ \cite{OSLSM}, and FSS-1000 \cite{fss}. PASCAL-$5^i$ and COCO-$20^i$ contain 20 and 80 object classes, respectively. And experiments on them are conducted in a cross-validation manner over four folds. For each fold, 1000 episodes are sampled for evaluation, while samples from the rest three folds are used for training. 
The FSS-1000 dataset containing 1000 classes is divided into 520, 240, and 240 classes for training, validation, and test.
We use two widely-adopted metrics~\cite{HSnet}: mean intersection over union (\text{mIoU}) and foreground-background IoU (FB-IoU).

\begin{figure*}[!ht]
\begin{minipage}{0.48\textwidth}
  \centering
  \includegraphics[width=0.95\linewidth]{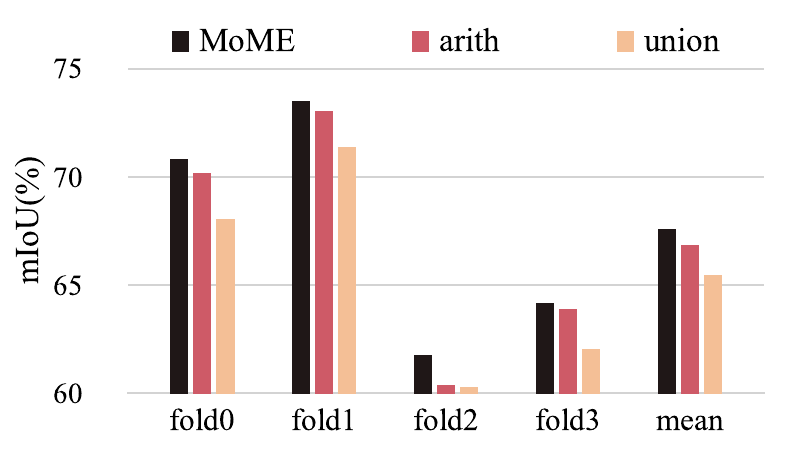}
  \vspace{-0.5em}
  \caption{Ablation study of \ourmodule module.}\vspace{-1.5em}
  \label{Fig:ablation_agg}
\end{minipage}
\hspace{1em}
\begin{minipage}{0.48\textwidth}
  \centering
  \includegraphics[width=0.9\linewidth]{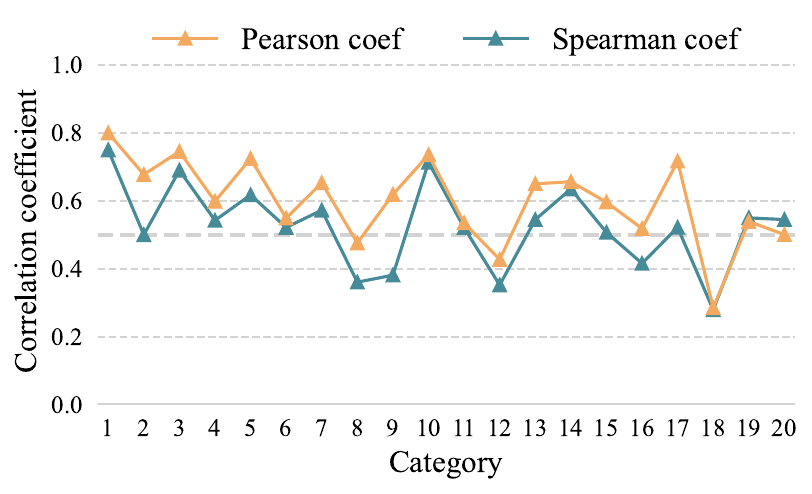}
  \vspace{-0.5em}
  \caption{Interpretability of the router in \ourmodule module.}\vspace{-1em}
  \label{Fig:analyse_agg}
\end{minipage}
\end{figure*}

\paragraph{Implementation Details}
We incorporate \ourapproach into two baseline models using the PyTorch framework, HSNet~\cite{HSnet} and ASNet~\cite{ASnet}, denoted as \ourapproach (HS) and \ourapproach (AS). We use 2 query images for training and 5 for testing, and add alignment regularization~\cite{PAnet} to facilitate  training. For backbone networks, we apply ResNet50 and ResNet101~\cite{resnet} and they are fixed in training. In addition, we experiment with the public pretrained model ViT-L/14@336px of CLIP~\cite{clip}, to evaluate the capability of our approach on the modern strong backbone. More details are in Appendix. 

\subsection{Comparison with State-of-the-Art Methods}
%定量分析
    %miou和fb-iou
    %FPS
    %domain shift
%定性分析：可视化结果
% \subsubsection{Quantitative Results}
% \label{sec:quantitative}
%coco/pascal/FSS
%FPS
In this subsection, we compare \ourapproach with multiple recent methods.
The results in Tab.~\ref{Tab:pascal_coco} (left) are on the COCO-20$^i$ dataset and provide the most noticeable gains. Particularly, \ourapproach (HS) and \ourapproach (AS) with ResNet101 obtain significant improvements of 7.5\% and 8.2\% on mIoU compared to the counterpart, respectively. 
Tab.~\ref{Tab:pascal_coco} (right) shows the results on the PASCAL-5$^i$ dataset. Our approach delivers consistent improvement (average over 3\% in mIoU). Tab.~\ref{Tab:fss} reports results of our approach in the 1-shot setting on the FSS-1000 dataset, which are all competitive with the counterparts in the 5-shot setting.

Furthermore, the bottom of Tab.~\ref{Tab:pascal_coco} shows results of additional experiments with CLIP on the two datasets. When using the strong backbone CLIP as the feature extractor, our \ourapproach approach still outperforms baseline models and recent methods under the 1-shot setting. In particular, \ourapproach (HS) obtain the improvements of 3.5\% in mIoU on the PASCAL-20$^i$ dataset. 
Such improvements further underscore the capability of our approach to effectively harness the diversity among queries and demonstrate its broad applicability to any backbone, even the most cutting-edge one. 
Furthermore, please see visualization results in Appendix.

\subsection{Ablation Study and Analysis}
We conduct extensive ablation experiments using the PASCAL-5$^i$ \cite{OSLSM} dataset and ResNet-50 \cite{resnet} backbone.

\subsubsection{Ablation Study on \ourmodule module}
How to effectively integrate candidate masks of the host query is crucial for fully unlocking the potential of supportive queries in mutual assistance. Thus we compare three alternative choices of candidate mask integration in Fig.~\ref{Fig:ablation_agg}. 
\textit{arith}: arithmetic mean scheme by calculating the average of all masks. \textit{union}: foreground union scheme by taking the union of foregrounds in all masks.
Obviously, the performance gaps between \textit{\ourmodule} and the others on four folds confirm the superiority of \ourmodule module.
\begin{table}[!ht]
    \caption{Performance on FSS-1000~\cite{fss} in 1-support $N_q$-query setting. ASNet$^5$: results in the conventional 5-support setting. $^\dagger$ASNet: Our reimplementation.}
    % based on official codes. 
    % We mark the best results in \textbf{bold}.}
    \vspace{-1em}
    \label{Tab:fss}
    \begin{center}
    \begin{small}
    \resizebox{0.9\linewidth}{!}{
    \begin{tabular}{c|l|c c}
        \toprule
        Setting & \multicolumn{1}{c|}{Method} & mIoU & FB-IoU \\
        \midrule
        %resnet50:
        \multicolumn{4}{c}{\textbf{\textit{Model performances with the ResNet-50 backbone}}}\\
        \midrule
        \multirow{1}{*}{5-support} 
        % & HSNet$^5$ \cite{HSnet} & \textit{87.5} & - \\
        & $^\dagger$ASNet$^5$ \cite{ASnet} & \textit{87.3} & \textit{92.1} \\
        \midrule
        \multirow{3}{*}{1-support} 
        % & SSP~\cite{SSP} & 87.3 & - \\
        % & Anno-free FSS~\cite{anno_free_FSS} & 85.7 & - \\
        & DifFSS~\cite{DiffFSS} & 88.4 & - \\
        % & HSNet \cite{HSnet} & 85.5 & - \\
        & $^\dagger$ASNet \cite{ASnet} & 86.0 & 91.0 \\
        \cmidrule{2-4}
         % & \ourapproach (HS) & \cellcolor{gray!20}88.0 & \cellcolor{gray!20}92.6 \\
        & \ourapproach (AS) & \cellcolor{gray!20}\textbf{88.2} & \cellcolor{gray!20}\textbf{92.6} \\
        \midrule
        %resnet101:
        \multicolumn{4}{c}{\textbf{\textit{Model performances with the ResNet-101 backbone}}}\\
        \midrule
        \multirow{1}{*}{5-support} 
        % & HSNet$^5$ \cite{HSnet} & \textit{88.5} & - \\
        & $^\dagger$ASNet$^5$ \cite{ASnet}& \textit{88.1} & \textit{92.6} \\
        \midrule
        \multirow{2}{*}{1-support} & 
        % SSP \cite{SSP} & 88.6 & - \\
        % & HSNet \cite{HSnet} & 86.5 & - \\
        $^\dagger$ASNet \cite{ASnet} & 87.0 & 91.7 \\
        \cmidrule{2-4}
        % & \ourapproach (HS) & \cellcolor{gray!20}88.2 & \cellcolor{gray!20}92.7 \\
        & \ourapproach (AS) & \cellcolor{gray!20}\textbf{88.9} & \cellcolor{gray!20}\textbf{93.1} \\
        \bottomrule
    \end{tabular}
    }
    \end{small}
    \end{center}
    \vspace{-2em}
\end{table}
\subsubsection{Interpretability of Scene-driven Router}
\begin{figure}[!ht]
    \centering
    \includegraphics[width=0.99\linewidth]{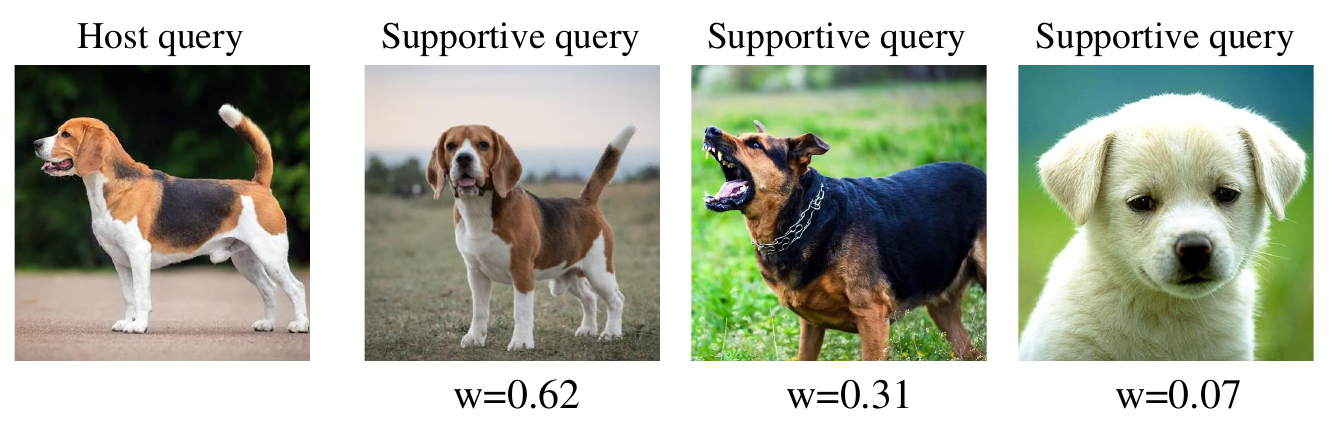}
    \vspace{-2em}
    \caption{Visualization results of different supportive queries and their corresponding router weights.}
    \vspace{-1em}
    \label{Fig:router_vis}
\end{figure}

We aim to enhance interpretability of the scene-driven router that guides the mixture of mask expert decisions in MoME module. As a widely-used metric in the field of style transfer, $\psi$ measures scene differences between two images (Calculation details in Appendix) and previous practice~\cite{BAM} has verified its effectiveness.
In this experiment, we sample 100 random image pairs from each category to compute the tried-and-true factor $\psi$ and router weights, and then calculate correlation coefficients between these two factors. In Fig.~\ref{Fig:analyse_agg}, Pearson and Spearman coefficients between $\psi$ and router weights are mostly above 0.5 (17/20 for Pearson and 15/20 for Spearman). Such phenomenon signals that $\psi$ and router weights are positively related, and the scene-driven router guides expert decisions based on scene differences between host and supportive queries. Some visualization results are in Fig.~\ref{Fig:router_vis}.

\subsubsection{Impacts of Query Amounts}
\begin{figure}[!ht]
    \centering
    \includegraphics[width=0.95\linewidth]{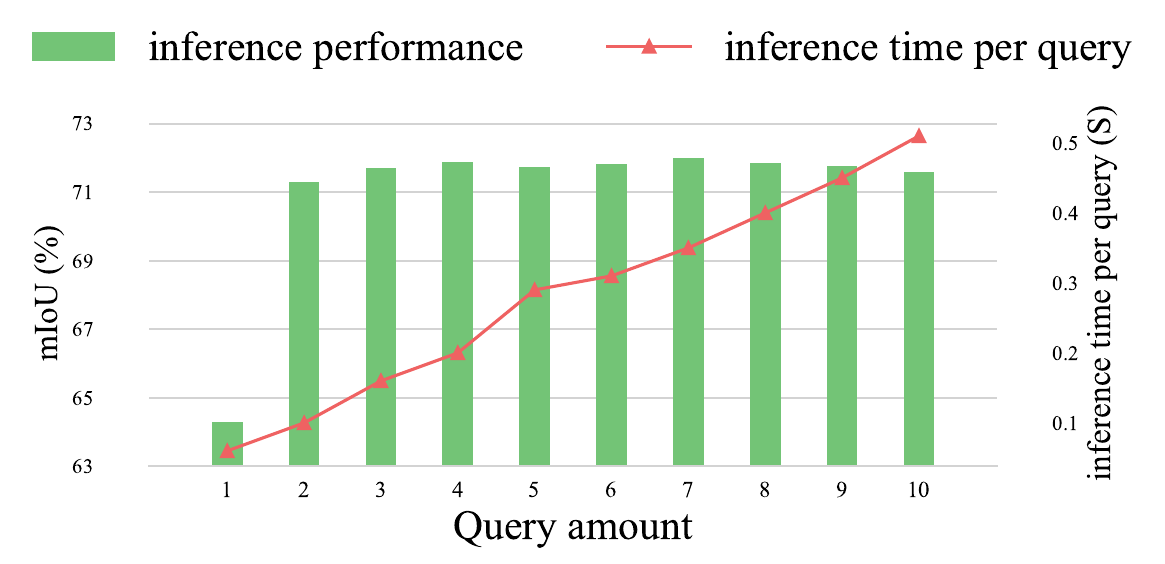}
    \vspace{-0.5em}
    \caption{Impacts of varying number of queries. We set the query amount to 2 given the balance of efficiency and performance.}
    \label{Fig:ablation_Nq}
    \vspace{-1em}
\end{figure}

The investigation of varying query amounts (\textit{i.e.}, mask expert amount) is in Fig.~\ref{Fig:ablation_Nq}.
As the number of query images increases, the inference performance first climbs up and then remains relatively stable. However, the inference time per query manifests an sustainable growth. Such trends indicate a marginal performance improvement at the cost of significant computational resources after reaching a point of stable fluctuation.
Therefore, one should select the most favorable query amount that balances both efficiency and performance, and in this paper we set the query amount to 2.

\section{Related Work}
We provide more detailed related works in this section. Few-shot segmentation is to make segmentation in the low-data regime. There are numerous methods, most of which investigated interactions between support and query images while the others focused on details inside either the support or query data.
\paragraph{Interactions between Support and Queries.}
%multi-scale/multi-layer
%transformer: cross-attention/not
%4D dimension
Many studies \cite{FECAnet2023, Buoys} sought to design high-capacity interaction modules between the support and query data. It was common to expand features in multiple scales and layers~\cite{BAM+multiscale} to enhance in-depth interplay. Based on the Transformer architecture, CyCTR~\cite{CyCTR}, DCAMA~\cite{DCAMA}, and HDMNet \cite{HDMnet2023} developed cross-attention operations while some others~\cite{VTM2023, MuHS2023, IPMT, AAFormer, IPRnet} did not. Besides, there were some attempts to generate high-dimensional feature correlation tensors, such as ASNet~\cite{ASnet}, VAT~\cite{VAT}. However, they tended to suffer from complex and expensive computations.

\paragraph{Support Exploration.}
%PAnet
%background: BAM, mine_latent
%boundary: MSI, HM
%label: MIAnet
%IPRnet: 批评它train with 5shot
Due to the scarcity of reference information, some studies took advantage of the support data in various perspectives (\textit{e.g.}, background and boundaries). PANet \cite{PAnet} proposed to take the predicted query data as pseudo references to segment support images in turn.
As the knowledge of objects from base classes in the background can be ample, it was investigated 
\cite{PMM, BAM} to use such knowledge to assist novel class predictions. HM \cite{HM} and MSI \cite{MSI} employed two masking techniques to segment more accurate target object boundaries. 
Recently, MIANet \cite{MIAnet2023} introduced supplementary category labels to leverage the general semantic information in a multi-modal manner.

\paragraph{Query Exploration.}
%列举
%没关注过query amount, 引出groupon
For query exploration, DPNet \cite{DPnet} added pseudo-prototypes extracted from the query foreground to the original support prototypes for refinement. SSP \cite{SSP} suggested that the locality of query masks, which were predicted in high confidence, could aid the support data to segment the leftover low-confidence areas. 
The known methods~\cite{QGnet} on query exploration mainly aimed to make full use of what is inside a single query image to guide the segmentation. In contrast, our \ourapproach approach exploits directly increasing the query amount to enrich semantic knowledge in an intuitive way. \ourapproach is more capable and simpler, and can be incorporated seamlessly with plentiful FSS baselines.

\section{Conclusions}
We proposed a novel and effective approach called \ourapproach to address the intra-class variation challenge in ONE-shot segmentation. 
The core innovation of \ourapproach is to leverage mutual information benefits of query images in a group for more representation multiplicities in a category. To make the most of multiple query images, we construct the \ourmodule module containing a scene-driven router and a flexible number of mask experts that jointly make comprehensive decisions. Experiments on three standard benchmarks showed significant performance enhancement and the effectiveness of our approach, which achieved comparable results with the counterparts under the 5-shot setting. In essence, we boosted one-shot segmentation from the data-centric perspective instead of the model architecture design and data labeling. We expect that our work will shed light on future research in addressing the intra-class variation issue.

\section*{Acknowledgment}

This research was partially supported by National Natural Science Foundation of China (No. 12326612), Zhejiang Key R\&D Program of China (No. 2023C03053), Zhejiang Key Laboratory of Medical Imaging Artificial Intelligence, State Key Laboratory of Transvascular Implantation Devices (NO. SKLTID2024003), and the Transvascular lmplantation Devices Research Institute (TIDRI).

\bibliographystyle{IEEEbib}
\bibliography{references}

\newpage
\appendix

\section{Experiments}
\subsection{Implementation Details.}
We provide more implementation details in this section. For backbone networks, we do not include VGG-16~\cite{vgg} due to its inferior performances compared with the ResNet family in the previous works, and instead we apply ResNet50~\cite{resnet} and ResNet101~\cite{resnet}, which are frozen to prevent them from learning class-specific representations of the training data~\cite{HSnet}. They have been pre-trained on ImageNet~\cite{imagenet} and are fixed to extract features (features from conv3\underline{\hspace{0.5em}}x to conv5\underline{\hspace{0.5em}}x before the ReLU \cite{RELU} activation of each layer are stacked to form the deep features). In the \ourmodule module, we select features from the 9th layer as input. The default settings for the optimizer and learning rate of the two baselines remain unchanged. Model training set to halt when it reaches the maximum 500th epoch. Note that our experiments demonstrate a wide coverage spanning across multiple datasets to various backbones in order to validate the effectiveness and generalizability of our \ourapproach approach, and they are done on four NVIDIA A100 Tensor Core GPUs. For evaluation metrics, mIoU averages over IoU values of all classes in the testing data to avoid the bias to the specific categories, while FB-IoU ignores object classes and computes average of foreground and background IoUs. As mIoU better reflects model generalization capability and prediction quality than FB-IoU does~\cite{HSnet}, we mainly focus on mIoU in our experiments.

\subsection{Evaluation Metrics.}
\label{metrics}
We evaluate the performance with two widely-adopted metrics~\cite{HSnet, PFEnet}: mean intersection over union (\text{mIoU}) and foreground-background IoU (FB-IoU), as:
\begin{equation}
\label{Eq:metric}
    \text{mIoU}=\frac{1}{n}\sum_{i=1}^n \text{IoU}_i,\ \ \ \ \ 
\end{equation}
\begin{equation}
\setlength{\abovedisplayskip}{1pt}
\setlength{\belowdisplayskip}{1pt}
\label{Eq:fbiou}
    \text{FB{\rm -}IoU}=\frac{1}{2}(\text{IoU}_\text{f} + \text{IoU}_\text{b}),
\end{equation}
where $\text{n}$ is the number of test samples, and $\text{f}$ means foreground and $\text{b}$ means background without regard to the object classes.

\subsection{Analysis of Datasets}
\label{App:datasets}
Due to more severe intra-class variation in COCO-20$^i$, \ourapproach achieves greater performance improvements in COCO-20$^i$ than that in PASCAL-5$^i$. In this section, we conduct analysis about the intra-class variation in these two datasets. 
There are overlapped 20 categories in COCO-20$^i$ and PASCAL-5$^i$, and we utilize the Affinity Propagation clustering algorithm within each category to cluster images based on image similarity. As represented in Tab.~\ref{Tab:cluster}, the average amount of clusters in COCO-20$^i$ is almost over four times than that in PASCAL-5$^i$. This further reflects the more severe intra-class variation issue in COCO-20$^i$ dataset from an intuitive perspective.
\begin{table*}[htbp]
\caption{The number of clusters in each category of PASCAL-5$^i$ and COCO-20$^i$ datasets.}
\label{Tab:cluster}
\setlength{\tabcolsep}{0.2em}
\resizebox{\textwidth}{!}{
\begin{tabular}{l|c|c|c|c|c|c|c|c|c|c|c|c|c|c|c|c|c|c|c|c|c}
\toprule
category & aeroplane & bicycle & bird & boat & bottle & bus & car & cat & chair & cow & diningtable & dog & horse & motorbike & person & pottedplant & sheep & sofa & train & tvmonitor & \textbf{avg} \\
\hline
pascal & 15 & 14 & 20 & 11 & 12 & 12 & 28 & 30 & 30 & 7 & 20 & 37 & 13 & 12 & 84 & 11 & 7 & 14 & 14 & 17 & \textbf{20.4} \\
\hline
coco & 59 & 51 & 42 & 42 & 82 & 67 & 127 & 68 & 159 & 33 & 148 & 63 & 57 & 64 & 514 & 64 & 35 & 88 & 62 & 96 & \textbf{96.1} \\
\bottomrule
\end{tabular}
}
\end{table*}
\begin{figure*}[htbp]
 \centering
 \subfloat{
     \begin{minipage}[c]{0.02\textwidth}
         \rotatebox{90}{\textbf{\small{support}}}
     \end{minipage}
     \hfill
     \begin{minipage}[c]{0.095\textwidth}
         \includegraphics[scale=0.115]{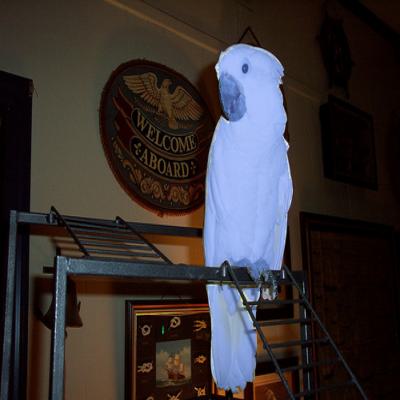}
     \end{minipage}
     \hfill
     \begin{minipage}[c]{0.095\textwidth}
         \includegraphics[scale=0.115]{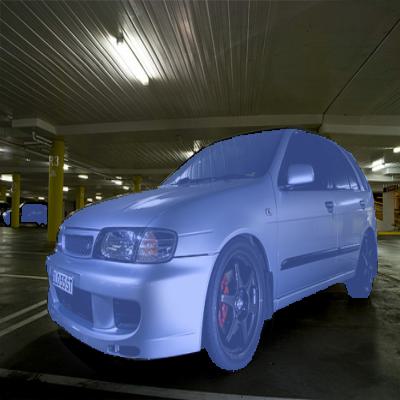}
     \end{minipage}
     \hfill
     \begin{minipage}[c]{0.095\textwidth}
         \includegraphics[scale=0.115]{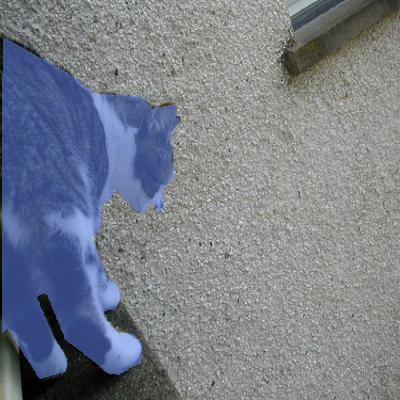}
     \end{minipage}
     \hfill
     \begin{minipage}[c]{0.095\textwidth}
         \includegraphics[scale=0.115]{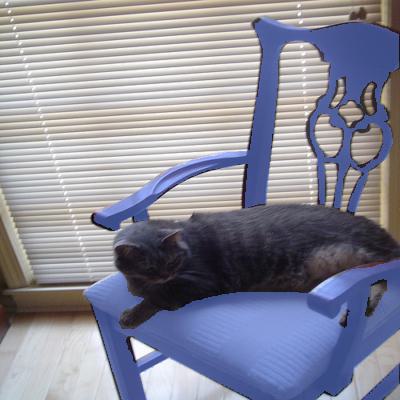}
     \end{minipage}
     \hfill
     \begin{minipage}[c]{0.095\textwidth}
         \includegraphics[scale=0.115]{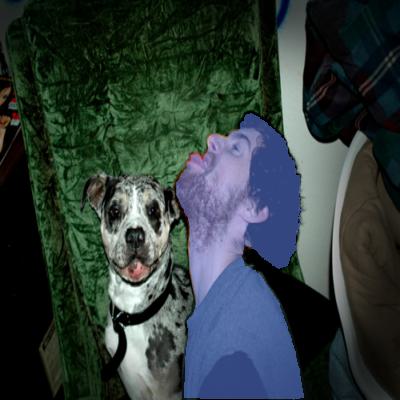}
     \end{minipage}
     \hfill
     \begin{minipage}[c]{0.095\textwidth}
         \includegraphics[scale=0.115]{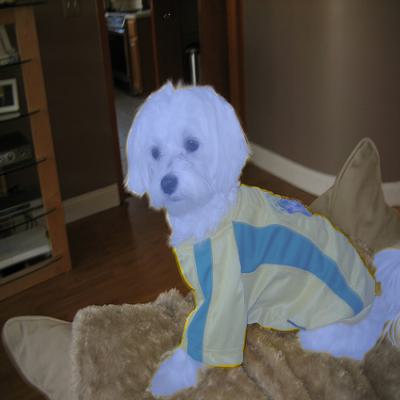}
     \end{minipage}
     \hfill
     \begin{minipage}[c]{0.095\textwidth}
         \includegraphics[scale=0.115]{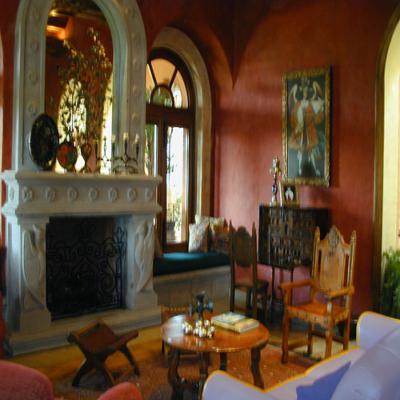}
     \end{minipage}
     \hfill
     \begin{minipage}[c]{0.095\textwidth}
         \includegraphics[scale=0.115]{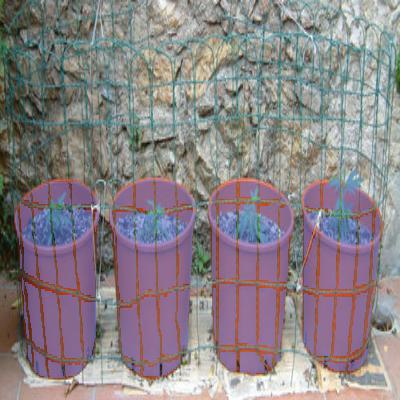}
     \end{minipage}
     \hfill
     \begin{minipage}[c]{0.095\textwidth}
         \includegraphics[scale=0.115]{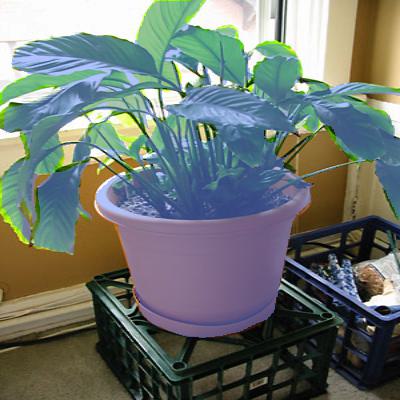}
     \end{minipage}
     % \hfill
     % \begin{minipage}[c]{0.095\textwidth}
     %     \includegraphics[scale=0.115]{vis10_s.jpg}
     % \end{minipage}
 }\vspace{1mm}
 \subfloat{
    \begin{minipage}[c]{0.02\textwidth}
         \rotatebox{90}{\textbf{\small{gt}}}
     \end{minipage}
     \hfill
     \begin{minipage}[c]{0.095\textwidth}
         \includegraphics[scale=0.115]{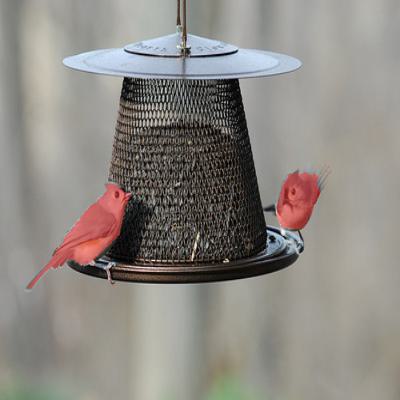}
     \end{minipage}
     \hfill
     \begin{minipage}[c]{0.095\textwidth}
         \includegraphics[scale=0.115]{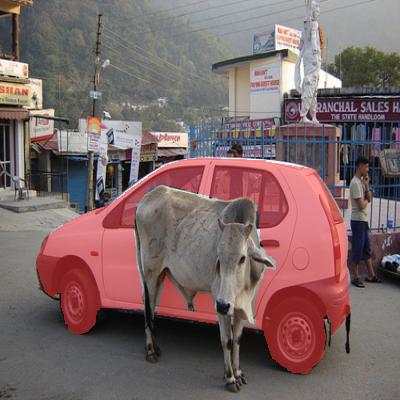}
     \end{minipage}
     \hfill
     \begin{minipage}[c]{0.095\textwidth}
         \includegraphics[scale=0.115]{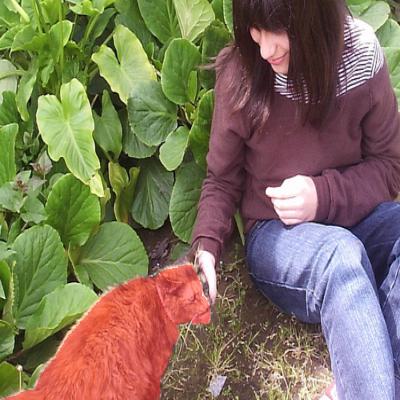}
     \end{minipage}
     \hfill
     \begin{minipage}[c]{0.095\textwidth}
         \includegraphics[scale=0.115]{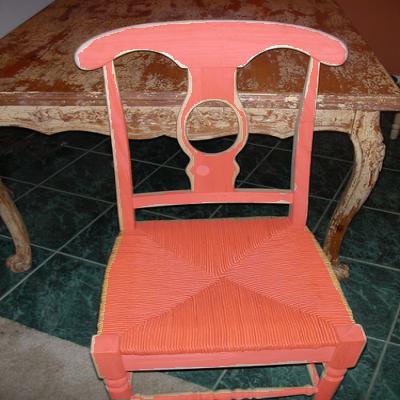}
     \end{minipage}
     \hfill
     \begin{minipage}[c]{0.095\textwidth}
         \includegraphics[scale=0.115]{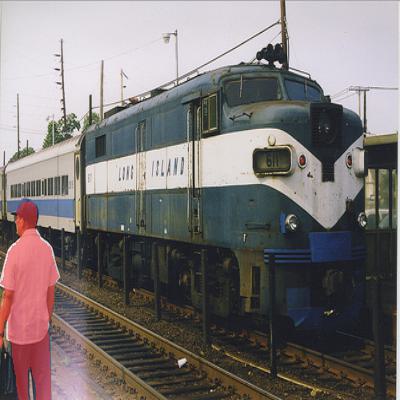}
     \end{minipage}
     \hfill
     \begin{minipage}[c]{0.095\textwidth}
         \includegraphics[scale=0.115]{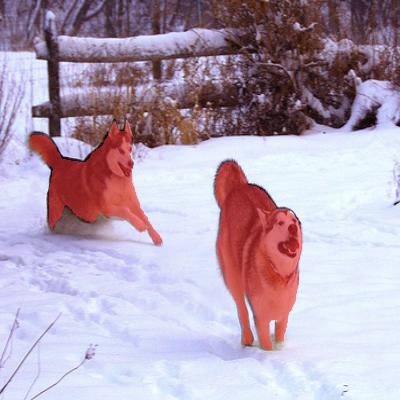}
     \end{minipage}
     \hfill
     \begin{minipage}[c]{0.095\textwidth}
         \includegraphics[scale=0.115]{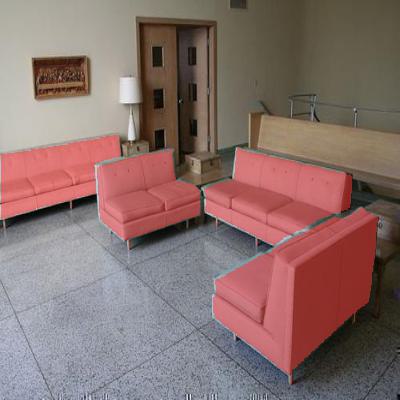}
     \end{minipage}
     \hfill
     \begin{minipage}[c]{0.095\textwidth}
         \includegraphics[scale=0.115]{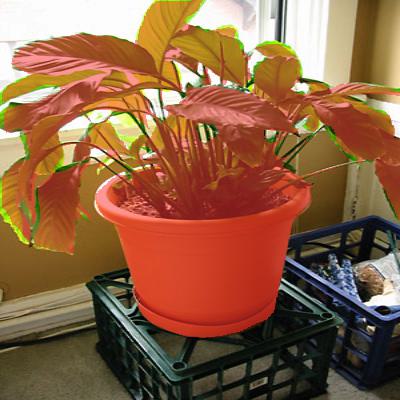}
     \end{minipage}
     \hfill
     \begin{minipage}[c]{0.095\textwidth}
         \includegraphics[scale=0.115]{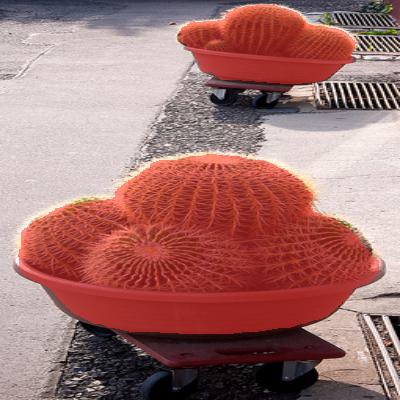}
     \end{minipage}
     % \hfill
     % \begin{minipage}[c]{0.095\textwidth}
     %     \includegraphics[scale=0.115]{vis10_gt.jpg}
     % \end{minipage}
 }\vspace{1mm}
 \subfloat{
     \begin{minipage}[c]{0.02\textwidth}
         \rotatebox{90}{\textbf{\small{baseline}}}
     \end{minipage}
     \hfill
     \begin{minipage}[c]{0.095\textwidth}
         \includegraphics[scale=0.115]{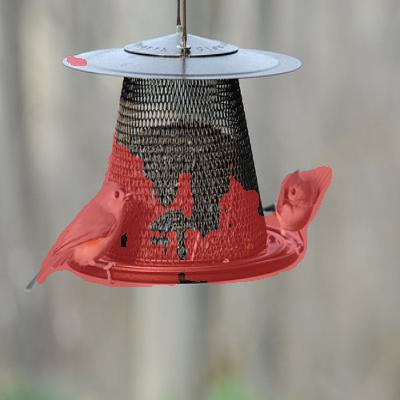}
     \end{minipage}
     \hfill
     \begin{minipage}[c]{0.095\textwidth}
         \includegraphics[scale=0.115]{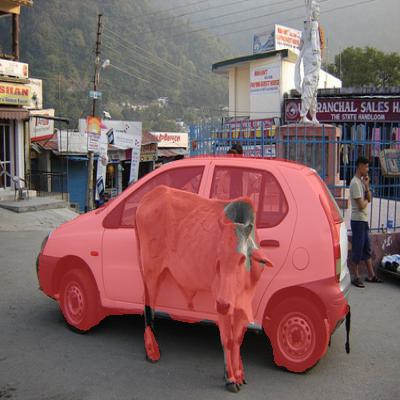}
     \end{minipage}
     \hfill
     \begin{minipage}[c]{0.095\textwidth}
         \includegraphics[scale=0.115]{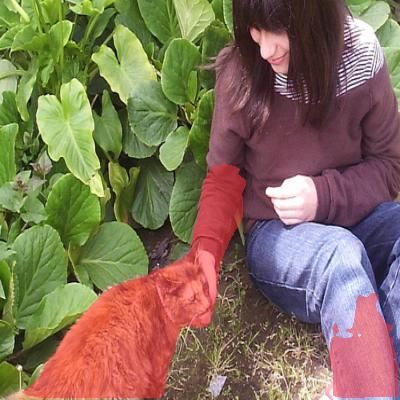}
     \end{minipage}
     \hfill
     \begin{minipage}[c]{0.095\textwidth}
         \includegraphics[scale=0.115]{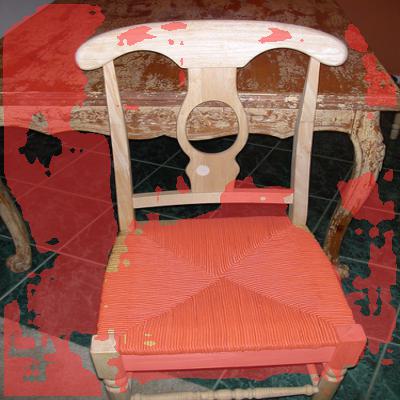}
     \end{minipage}
     \hfill
     \begin{minipage}[c]{0.095\textwidth}
         \includegraphics[scale=0.115]{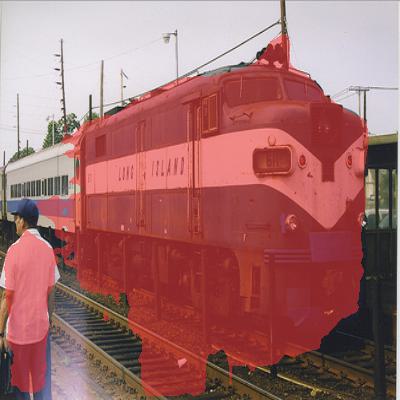}
     \end{minipage}
     \hfill
     \begin{minipage}[c]{0.095\textwidth}
         \includegraphics[scale=0.115]{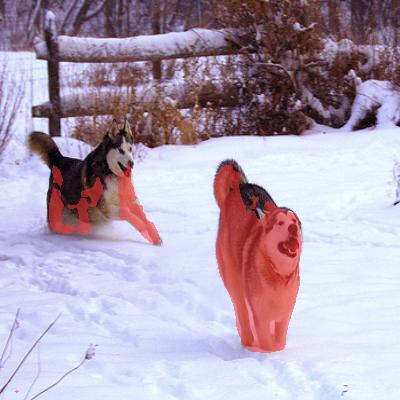}
     \end{minipage}
     \hfill
     \begin{minipage}[c]{0.095\textwidth}
         \includegraphics[scale=0.115]{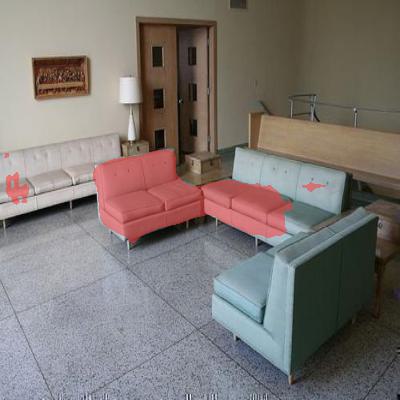}
     \end{minipage}
     \hfill
     \begin{minipage}[c]{0.095\textwidth}
         \includegraphics[scale=0.115]{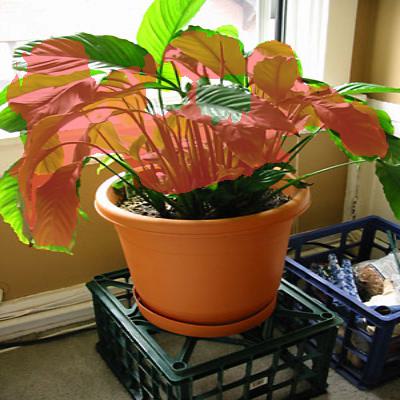}
     \end{minipage}
     \hfill
     \begin{minipage}[c]{0.095\textwidth}
         \includegraphics[scale=0.115]{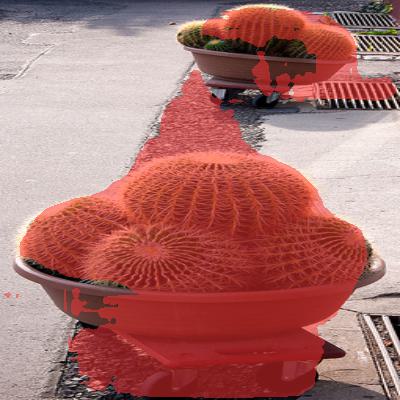}
     \end{minipage}
     % \hfill
     % \begin{minipage}[c]{0.095\textwidth}
     %     \includegraphics[scale=0.115]{vis10_hs.jpg}
     % \end{minipage}
 }\vspace{1mm}
 \subfloat{
    \hspace{0.2mm}
     \begin{minipage}[c]{0.02\textwidth}
         \rotatebox{90}{\textbf{\small{ours}}}
     \end{minipage}
     \hspace{-2mm}
     \begin{minipage}[c]{0.095\textwidth}
         \includegraphics[scale=0.115]{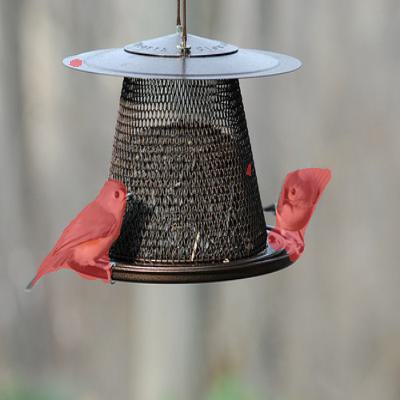}
     \end{minipage}
     \hfill
     \begin{minipage}[c]{0.095\textwidth}
         \includegraphics[scale=0.115]{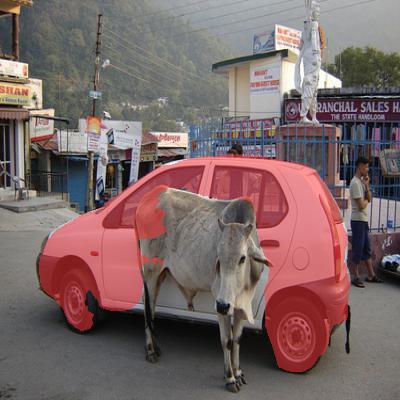}
     \end{minipage}
     \hfill
     \begin{minipage}[c]{0.095\textwidth}
         \includegraphics[scale=0.115]{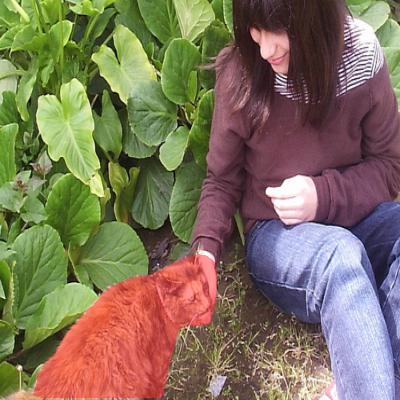}
     \end{minipage}
     \hfill
     \begin{minipage}[c]{0.095\textwidth}
         \includegraphics[scale=0.115]{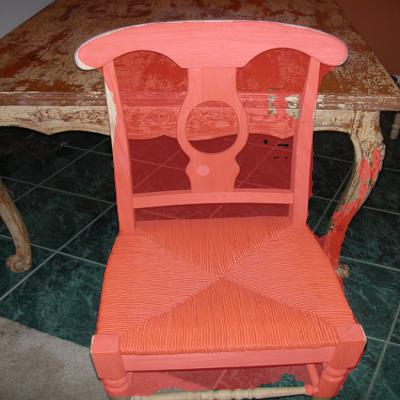}
     \end{minipage}
     \hfill
     \begin{minipage}[c]{0.095\textwidth}
         \includegraphics[scale=0.115]{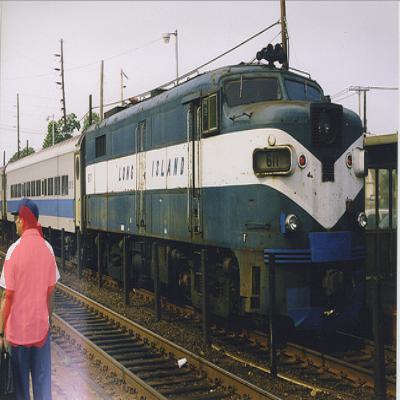}
     \end{minipage}
     \hfill
     \begin{minipage}[c]{0.095\textwidth}
         \includegraphics[scale=0.115]{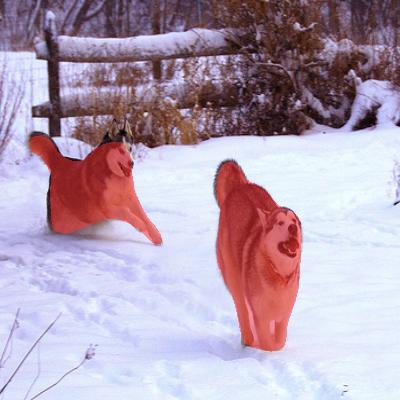}
     \end{minipage}
     \hfill
     \begin{minipage}[c]{0.095\textwidth}
         \includegraphics[scale=0.115]{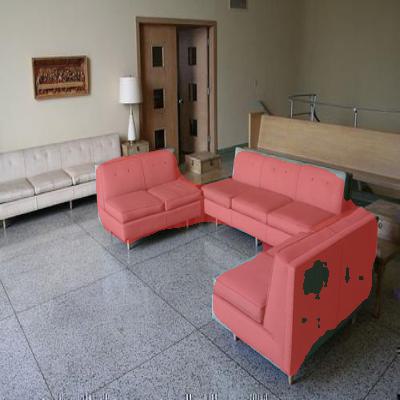}
     \end{minipage}
     \hfill
     \begin{minipage}[c]{0.095\textwidth}
         \includegraphics[scale=0.115]{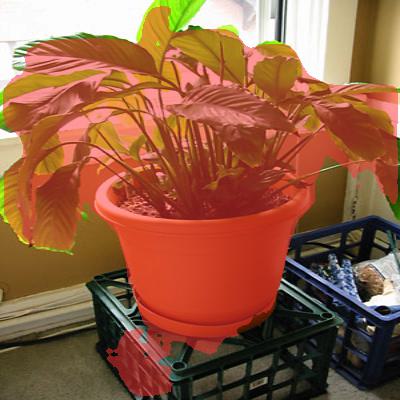}
     \end{minipage}
     \hfill
     \begin{minipage}[c]{0.095\textwidth}
         \includegraphics[scale=0.115]{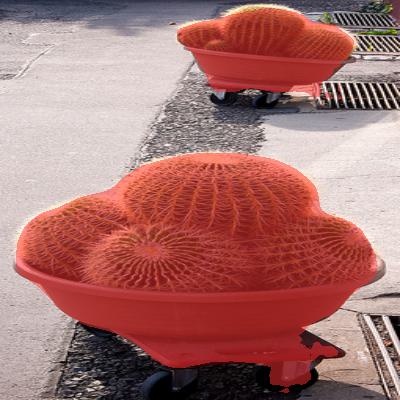}
     \end{minipage}
     % \hfill
     % \begin{minipage}[c]{0.095\textwidth}
     %     \includegraphics[scale=0.115]{vis10_ours.jpg}
     % \end{minipage}
 }
 \caption{Some qualitative results on the PASCAL-5$^i$~\cite{pascal} dataset in the one-shot setting, in the presence of intra-class variation, size differences, duplicated backgrounds, and occlusions.}
 \label{Fig:vis}
 \vspace{-1em}
\end{figure*}
\subsection{Interpretability of Router Weights in \ourmodule}
\label{App:voting}
First, the low-level feature maps $\textit{F}_{\text{low}}^{\text{host}},~\textit{F}_{\text{low}}^{\text{supp}} \in \mathbb{R}^{H_1 \times W_1 \times C_1}$ of the host query and supportive query are extracted from the 3th layer of the total features. Then we reshape them and calculate the Gram Matrices, following the previous work\cite{BAM}. Please note that the relevant operations of these two queries are similar, and that of the host one can be summarized as:
\begin{equation}
    \textit{A}_{\text{host}}=\mathcal{F}_{\text{reshape}}(\textit{F}_{\text{low}}^{\text{host}}) \in \mathbb{R}^{C_1 \times N_1},
\end{equation}
\begin{equation}
    \textit{G}^{\text{host}}=\textit{A}_{\text{host}}\textit{A}_{\text{host}}^\mathsf{T} \in \mathbb{R}^{C_1 \times C_1},
\end{equation}
where $N_1=H_1 \times W_1$ and $\mathcal{F}_{\text{reshape}}$ reshapes the size of input tensor to $C_1 \times N_1$. With the calculated Gram Matrices, the Frobenius norm is evaluated on their difference to obtain the overall indicator $\psi$:
\begin{equation}
    \psi=\Vert \textit{G}^{\text{host}}-\textit{G}^{\text{supp}} \Vert_{F},
\label{Eq:psi}
\end{equation}
where $\Vert \cdot \Vert_{F}$ denotes the Frobenius norm of the input matrix.

\subsection{Qualitative Results}
To intuitively display our approach, we visualize some results containing challenging intra-class variation under the one-shot setting in Fig.~\ref{Fig:vis}. 
The first row is support images with their masks in blue, while the second row represents the groundtruth. And the following two rows display predictions of the baseline model HSNet~\cite{HSnet} and our approach, respectively. 
Note that all the images in each column belong to the same category. One can see from our results (the 4th row) that if objects in the support image differ greatly from those in the query images, our \ourapproach approach captures the semantic essence of the objects and segments more accurately with the reciprocal knowledge help of the multiple query images.

\end{document}